\documentclass{article}
\usepackage{ijcai22}

\usepackage{times}
\usepackage{soul}
\usepackage{url}

\usepackage[dvipsnames]{xcolor}
\usepackage[hidelinks,colorlinks,citecolor=MidnightBlue]{hyperref}
\usepackage[utf8]{inputenc}
\usepackage[small]{caption}
\usepackage{graphicx}
\usepackage{amsmath}
\usepackage{amsthm}
\usepackage{booktabs}
\usepackage{algorithm}
\usepackage{algorithmic}
\usepackage{amssymb}
\usepackage{color}
\usepackage{subfigure}
\usepackage{graphicx}
\usepackage{multirow}
\usepackage{marvosym}
\usepackage{ifsym}
\usepackage[colorlinks,linkcolor=red]{}

\DeclareMathOperator*{\argmax}{argmax}

\title{Zero-Shot Logit Adjustment}

\author{
	Dubing Chen$^{1*}$
	\and
	Yuming Shen$^{2*}$\and
	Haofeng Zhang\textsuperscript{$1$\Letter}\and
	Philip H.S. Torr$^2$
	\affiliations
	$^1$Nanjing University of Science and Technology\\
	$^2$University of Oxford\\
	\emails
	\{db.chen, zhanghf\}@njust.edu.cn,
	ymcidence@gmail.com,
	philip.torr@eng.ox.ac.uk
}

\begin{document}

	\maketitle
	
	\begin{abstract}
		Semantic-descriptor-based Generalized Zero-Shot Learning (GZSL) poses challenges in recognizing novel classes in the test phase. The development of generative models enables current GZSL techniques to probe further into the semantic-visual link, culminating in a two-stage form that includes a generator and a classifier. However, existing generation-based methods focus on enhancing the generator's effect while neglecting the improvement of the classifier. In this paper, we first analyze of two properties of the generated pseudo unseen samples: bias and homogeneity. Then, we perform variational Bayesian inference to back-derive the evaluation metrics, which reflects the balance of the seen and unseen classes. As a consequence of our derivation, the aforementioned two properties are incorporated into the classifier training as seen-unseen priors via logit adjustment. The Zero-Shot Logit Adjustment further puts semantic-based classifiers into effect in generation-based GZSL. Our experiments demonstrate that the proposed technique achieves state-of-the-art when combined with the basic generator, and it can improve various generative Zero-Shot Learning frameworks. Our codes are available on \url{https://github.com/cdb342/IJCAI-2022-ZLA}.
	\end{abstract}
		\section{Introduction}

	In recognition tasks, it is challenging when classes for training and test are different, known as Zero-Shot Learning (ZSL) problems, which call for knowledge generalization from seen to unseen classes. The goal of ZSL is to correctly recognize unseen samples with a classifier trained on seen classes. To bridge the gap between training and test domains, it counts on the similarity of semantic descriptors \cite{lampert2009learning} corresponding to each class.\let\thefootnote\relax\footnotetext{* Equal contribution.} 
        \let\thefootnote\relax\footnotetext{\Letter \ Corresponding author. This work was supported by the National Natural Science Foundation of China (NSFC) under Grants No. 61872187, No. 62077023 and No. 62072246.}Through the semantic descriptors, ZSL can transfer knowledge from seen to unseen classes without accessing the data of unseen classes.

	Early ZSL works focus on end-to-end classifier learning, particularly the embedding models \cite{lampert2013attribute,li2019rethinking} which match visual features and semantic descriptors in a shared embedding space. These methods appear effective in ZSL but perform poorly in the more challenging Generalized Zero-Shot Learning (GZSL) setting \cite{chao2016empirical,xian2017zero}. More recent efforts \cite{xian2018feature,shen2020invertible} aim to improve GZSL performance by decomposing the learning process into generator learning and classifier learning. Such a two-stage strategy compensates for the feature expression of unseen classes during the classifier learning by the generated samples. A typical family of studies has focused on the improvement of the generator, exploring alternative architectures, or introducing various inductive biases. On the contrary, there is a scarcity of research on classifiers under the generative paradigm.
		\begin{figure}[t]
		\centering
\includegraphics[width=0.48\textwidth]{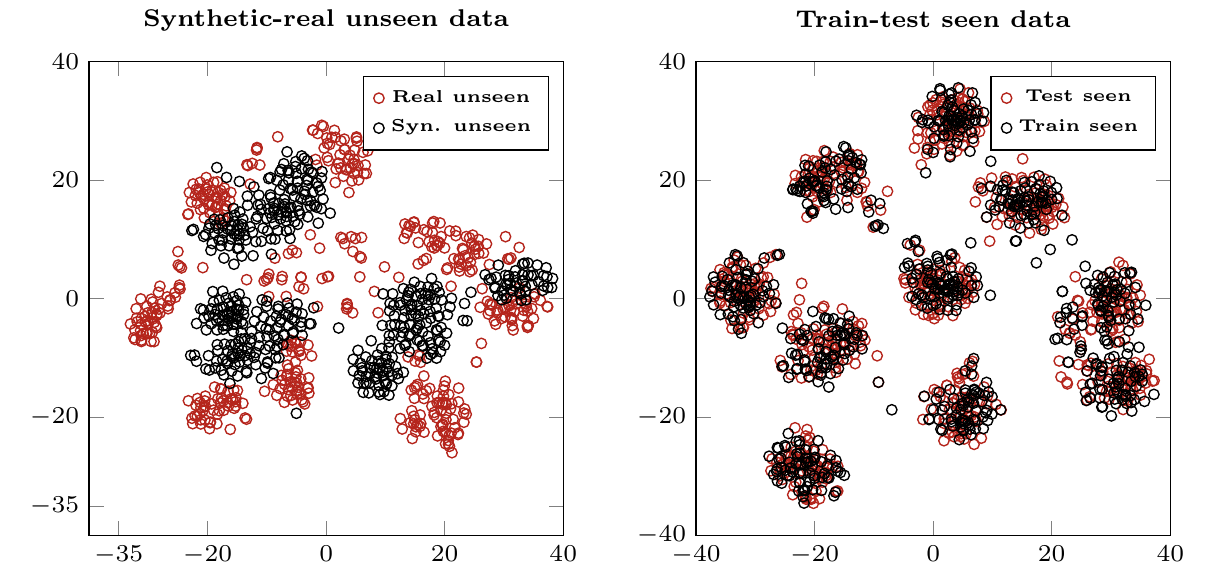}
		\vspace{-1.5ex}
		\caption{t-SNE visualization of the synthetic-real unseen data (left) and the real train-test seen data (right) in AWA2, we sample the same number for each class pair. The synthetic unseen data shows obvious domain bias from real test unseen data, while the training domain of the seen classes is consistent with the test domain.}
		\vspace{-2ex}
		\label{fig1}
	\end{figure}
    
	A principled classifier design is required to uncap the performance of the two-stage approach. The intuitive notion is introducing semantic information to classifier learning. However, both semantic and visual embedding strategies learn the same knowledge as generative models, i.e., semantic-visual links. It implies that, by directly replacing the classifier, the generated unseen samples can only serve to deliver the generator's learned links to the classifier, with little effect. As a result, classic ZSL classifiers perform worse than vanilla softmax classifiers in the generative setting \cite{xian2018feature}.

	To determine design directions for the classifier, we first investigate the distribution of synthetic unseen data. As shown in Figure \ref{fig1}, the synthetic unseen samples deviate severely from the real distribution. This bias is theoretically inevitable since the domain shift problem \cite{fu2014transductive}, which leads to the misalignment of training and test domains. Nevertheless, we further explore the working mechanism of generated samples in Table \ref{tab1}, finding that the information contained in various synthetic unseen samples is quite homogenous for the classifier. Existing methods employ a large sampling size from generated unseen distribution to magnify the influence of modest valid information included in synthetic unseen samples on feature expression. However, such a resampling strategy has been proven to result in overfitting of the classifier on certain feature patterns (the biased generated unseen distribution in this setting) in other fields \cite{buda2018systematic,menon2020long}, which harms the recognition of both seen and unseen classes.
	
	These findings imply that there is a need to restrict the generation number of unseen samples. Can we directly transmit class imbalance information to the classifier during training rather than inefficiently increasing the expression of unseen classes by resampling? In this work, we regard the \textbf{bias} and \textbf{homogeneity} of generated unseen samples as special prior knowledge. In light of the success of loss modification in class imbalance research \cite{lin2017focal,menon2020long}, we incorporate this seen-unseen prior into the classifier training process in the form of logit adjustment. Specifically, we establish the lower bound of the seen-unseen balanced accuracy \cite{xian2017zero} with variational Bayesian inference and obtain an adjusted posterior. Then the prior takes part in training via adjusting the logits of vanilla cross-entropy loss. This approach, termed Zero-Shot Logit Adjustment (ZLA), allows for a lower number of generated unseen samples while producing more balanced results. By establishing a semantic-prototype mapping, we further introduce the semantic information to the classifier. Notably, our proposed ZLA allows the generated unseen samples to play an adjustment role rather than supplying unseen class information, which overcomes the embedding-based classifier's previous ineffectiveness in the generative scenario (see Sec. \ref{sec4.5}). Our contributions are summarized as follows:
	
	\begin{itemize}
\item We mathematically derive the lower bound of the seen-unseen balanced accuracy, allowing us to include generated unseen samples's bias and homogeneity as a seen-unseen prior in cross-entropy via logit adjustment.
\item Based on ZLA, we break the previous classifier's inconsistency in training objectives and test metrics and the inability to exploit semantic priors in classifier learning.
\item Our proposed classifier enables greatly reducing the generation number of unseen samples. It outperforms SoTAs when combined with the base generator, and can be a plug-in to augment various generation-based methods.
	\end{itemize}
	
	\section{Related Work}
	\subsection{Zero-Shot Learning}
Zero-Shot Learning (ZSL) \cite{lampert2009learning,farhadi2009describing} has become a popular research area in recent years. Classic ZSL excludes
seen classes during the test, while Generalized Zero-Shot Learning (GZSL) \cite{chao2016empirical,xian2017zero} considers both seen and unseen classes, attracting more current interest. The typical embedding-based ZSL methods \cite{li2019rethinking,skorokhodov2020class} learn the semantic-visual links for classification, but with little effectiveness in the GZSL scenario. The progress of GZSL was once driven by the development of generative models \cite{kingma2013auto,goodfellow2014generative}, which allowed converting the GZSL problem to common supervised classification using a generator-classifier architecture. Until recently, research on generators \cite{shen2020invertible,han2021contrastive} gradually saturated, whereas classifier design is rarely examined in such a two-stage framework. To break the bottleneck of generation-based approaches, the principle design of a classifier is required.

\subsection{Posterior Modification}
Posterior modification, which has been deeply studied in class imbalance learning \cite{lin2017focal,menon2020long}, aims at producing a class-balanced prediction. Post-hoc correction \cite{collell2016reviving}, loss re-weighting\cite{menon2013statistical}, and logit adjustment \cite{menon2020long} are its representative strategies. A similar procedure has been adopted in certain ZSL research. DCN \cite{liu2018generalized} utilizes entropy regularization to calibrate the predictions of seen and unseen classes. The post-hoc correction, known as calibrated stacking \cite{chao2016empirical} in ZSL, is also employed. However, a more general strategy in generation-based GZSL is to sample a large number of unseen class samples from the generated distribution. Although the re-sampling technique \cite{chawla2002smote} has been proven to produce overfitting in long-tail learning \cite{collell2016reviving}, its shortcoming in generation-based GZSL is a lack of exploration. In this paper, we mathematically introduce the more advanced logit adjustment strategy into GZSL for a better balance between seen and unseen predictions.

	\section{Methodology}
	
	Considering two disjoint label sets, $\mathcal{Y}^{s}$ and $\mathcal{Y}^{u}$, GZSL aims at recognizing instances that belong to $\mathcal{Y}^{s}\cup\mathcal{Y}^u$, while only accessing samples with labels in $\mathcal{Y}^{s}$ during training. Define the visual space $\mathcal{X} \subseteq \mathbb{R}^{d_x}$ and the semantic set $\mathcal{A}=\{\mathbf{a}_y|y\in \mathcal{Y}^{s}\cup\mathcal{Y}^u\} \subseteq \mathbb{R}^{d_a}$, where $d_x$ and $d_a$ are dimensions of these two spaces. Then the goal of GZSL is to learn such a classifier, i.e., $f_{gzsl}:\mathcal{X}\rightarrow\mathcal{Y}^{s}\cup\mathcal{Y}^u$ given the training set $\mathcal{D}^{tr}=\{ \mathbf{x},y |\mathbf{x}\in \mathcal{X},y\in \mathcal{Y}^{s}\}$ and the global semantic set $\mathcal{A}$.
	
	The two-stage framework typically processes this problem with two main components: the generator $\mathit{G}$ and the classifier $\mathit{C}$. The generator $\mathit{G}$, defined as an arbitrary generative model \cite{kingma2013auto,goodfellow2014generative}, is first trained with the seen visual features and their corresponding semantics for conditionally mapping the Gaussian noise to fit the real visual distribution. The instances generated by unseen class semantics and the real seen instances are then fed into the classifier $\mathit{C}$ together to fit the posterior probability:
	\begin{equation}
	\begin{split}
    \begin{aligned}
	\mathbf{\hat{x}}=\mathit{G}(\mathbf{z},\mathbf{a}),&\mathbf{z}\sim\mathcal{N}(\mathbf{0},\mathbf{I}),\\ p(y|\mathbf{\widetilde{x}}):=\mathrm{softmax}[&\mathit{C}(\mathbf{\mathbf{\widetilde{x}}})], \mathbf{\widetilde{x}}\in \mathcal{X} \cup \{\mathbf{\hat{x}}\},
    \end{aligned}
	\end{split}
    \label{eq1}
	\end{equation}
	where $\mathbf{\hat{x}}$ denotes the synthetic instances, $\mathbf{z}$ represents random Gaussian noises, and $\mathbf{\widetilde{x}}$ is either real or synthetic instances. $p(y|\mathbf{\widetilde{x}})$ denotes the posterior probability derived from the classifier. Then the model predicts the class label by taking $\mathcal{C}_{out} = {\argmax}_y(p(y|\mathbf{\widetilde{x}}))$. In this work, we focus on the design of the classifier under the GZSL setting.
	
	\subsection{Preliminary: Logit Adjustment }
	
	Logit adjustment strategy is commonly employed in class imbalance tasks \cite{lin2017focal,menon2020long}, which weights the logit in softmax cross-entropy, i.e., 
	\begin{equation}
	\resizebox{.91\linewidth}{!}{$
		\displaystyle
		\mathcal{L}_{LA}=\mathrm{log}[1+\sum_{y' \ne y}\delta(y,y')  \mathrm {exp}(\mathcal{C}_{y'}(x)-\mathcal{C}_y(x))],
		$}
        \label{eq2}
	\end{equation}
	where $\mathcal{C}_{y'}(\cdot)$ is the logit corresponding to class $y'$, and $\delta(y,y')$ represents the adjustment weight. The larger $\delta(y,y')$ results in the network focusing more on optimizing the logit of $y'$, allowing control of the prediction probabilities of different categories. Existing class imbalance works typically associate the weights with the class prior of $y$ or $y'$ \cite{cao2019learning,tan2020equalization,menon2020long}.
	
	\subsection{Empirical Analysis on Generated Samples }
\label{prior}
	\begin{table}[H]
		\centering
        	\resizebox{0.33\textwidth}{!}{
		\begin{tabular}{ccccccc}
			\toprule
			Method                & G.N.          &T1   & $\mathcal{A}^U$    & $\mathcal{A}^S$    & $\mathcal{A}^H$    \\
			\midrule
			\multirow{2}{*}{MSE}  & 588     & 67.9 & 17.1 & 71.6 & 27.6 \\
			 & 4000                      &67.5 & 57.1 & 59.7 & 58.4 \\
			\multirow{2}{*}{VAE}    & 588                  &68.0 & 25.8 & 67.6 & 37.3 \\
		 & 4000                           &68.6 & 57.2 & 68.8 & 62.9 \\
			\multirow{2}{*}{WGAN}   & 588                          &68.7 & 20.9 & 83.2 & 33.5 \\
			 & 4000                    &68.3   & 57.7 & 71.0 & 63.7\\
			\bottomrule
		\end{tabular}
        }
		\caption{ZSL (T1) and GZSL ($\mathcal{A}^H$) results of the simple semantic-visual mapping net (denoted as MSE) and two different generative models, VAE and WGAN on AWA2. {\bf G.N.} denotes the generation number per unseen class (588 is the class averaged number of real seen samples).}
        \vspace{-2ex}
		\label{tab1}
	\end{table}
	Regardless the bias problem (Figure \ref{fig1}) of generated unseen samples, generation-based methods achieve a certain success in GZSL. Thus, we empirically investigate the working mechanism of the generated unseen samples. As shown in Table \ref{tab1}, we compare the single class center point (semantic to visual-center mapping trained with MSE loss) resampling technique with two generative models, i.e., VAE \cite{kingma2013auto} and WGAN \cite{gulrajani2017improved} (detailed in supplementary material). Two phenomena can be intuitively observed by comparing the results in Table \ref{tab1}: (1) the key success of generation-based models in GZSL relies on unseen class feature expression enhancement by a large number of generated unseen samples; and (2) the more diversified samples (generated from generative models) produce a limited performance improvement compared to replicating a single point. We forego a deeper study due to its orthogonal nature to our work, but these modest findings imply that the synthetic unseen samples are rather homogenous than real ones.
	
	Despite the difficulty of eliminating bias and homogeneity, can we include them in classifier training as the seen-unseen prior? Below, we'll illustrate how, by changing the classifier's optimization target, we can integrate this prior into the learning process in the form of the logit adjustment.
	\subsection{From the Statistical View}
The nature of GZSL is an extreme case of class imbalance, as measured by a class balanced metric $\mathcal{A}^H$ (detailed in Sec. \ref{metric}). Assuming the class space has been completed by a set of pseudo unseen class samples generated by the trained generator, existing classifiers optimize the global accuracy $\mathcal{A}^G$ by modeling the base posterior probability (Eq. \ref{eq1}):
\vspace{-0.3ex}
    \begin{equation}
    \mathcal{A}^G=\mathbb{E}_{\mathbf{x}\sim p(\mathbf{x})} q(\mathcal{C}_{out}=y_{\mathbf{x}}|\mathbf{x}),
    \label{ag}
    \end{equation}
where $p(\mathbf{x})$ is defined as a uniform distribution over all data, $y_{\mathbf{x}}$ is the true label of $\mathbf{x}$, and $q(\mathcal{C}_{out}=y|\mathbf{x})$ is the probability to predict class $y$ with the classifier $\mathcal{C}$. However, Eq. \ref{ag} neglects the imbalance between seen and unseen domains, which is inconsistent with the test metric $\mathcal{A}^H$. To find balanced results across classes, we in turn employ evaluation indicators to guide the design of the classifier. Indeed, let $\mathcal{A}(y)$ denote the accuracy in class $y$, we have \cite{collell2016reviving}
\vspace{-0.5ex}
	\begin{equation}
	\mathcal{A}(y)=\mathbb{E}_{\mathbf{x}\sim p(\mathbf{x})}\frac{ q(\mathcal{C}_{out}=y|\mathbf{x})p(y|\mathbf{x})}{p(y)},
    \label{eq3}
	\end{equation}
	where $p(y)$ represents the statistics frequency of class $y$, and $p(y|\mathbf{x})$ denotes the real posterior probability. Then the average accuracy of seen classes is
    \vspace{-0.2ex}
	\begin{equation}
	\begin{aligned}
	\mathcal{A}^S&=\frac{1}{|\mathcal{Y}^s|} \sum_{y \in \mathcal{Y}^s} \mathcal{A}(y)\\
	&=\frac{1}{|\mathcal{Y}^s|} \sum_{y \in \mathcal{Y}^s}\mathbb{E}_{\mathbf{x}\sim p(\mathbf{x})}\frac{ q(\mathcal{C}_{out}=y|\mathbf{x})p(y|\mathbf{x})}{p(y)}\\
	&= \mathbb{E}_{y\in \mathcal{Y}^s}\mathbb{E}_{\mathbf{x}\sim p(\mathbf{x})}\frac{ q(\mathcal{C}_{out}=y|\mathbf{x})p(y|\mathbf{x})}{p(\mathcal{Y}^s)p(y|y\in \mathcal{Y}^s)},
	\end{aligned}
    \label{eq4}
	\end{equation}
	where $p(y|y\in \mathcal{Y}^s)$ denotes the frequency of class $y$ in $\mathcal{Y}^s$, and $p(\mathcal{Y}^s)$ is a theoretically derived data-independent probability which will be served as a hyperparameter. Analogously, the average accuracy of unseen classes is
    \vspace{-0.2ex}
	\begin{equation}
	\begin{aligned}
	\mathcal{A}^U= \mathbb{E}_{y\in \mathcal{Y}^u}\mathbb{E}_{\mathbf{x}\sim p(\mathbf{x})}\frac{ q(\mathcal{C}_{out}=y|\mathbf{x})p(y|\mathbf{x})}{p(\mathcal{Y}^u)p(y|y\in \mathcal{Y}^u)}.
	\end{aligned}
    \label{eq5}
	\end{equation}
    
Then we consider the harmonic mean, $\mathcal{A}^H$, which serves the target of attaining high accuracy for both seen and unseen classes, and empirically reaches its maximum when the accuracy of seen and unseen classes is balanced \cite{xian2017zero}. We have
\vspace{-1.7ex}
	\begin{equation}
	\displaystyle
	\mathcal{A}^H=2/(\frac{1}{\mathcal{A}^S}+\frac{1}{\mathcal{A}^U}).
    \label{eq6}
	\end{equation}
    
	Based on the convex property of the inversely proportional function, we have the upper bound of $1/{\mathcal{A}^S}$ with Jensen Inequality:
    \vspace{-0.5ex}
	\begin{equation}
	\begin{aligned}
	\displaystyle
	\frac{1}{\mathcal{A}^S}&=1/\mathbb{E}_{y\in \mathcal{Y}^s}\mathbb{E}_{\mathbf{x}\sim p(\mathbf{x})}\frac{ q(\mathcal{C}_{out}=y|\mathbf{x})p(y|\mathbf{x})}{p(\mathcal{Y}^s)p(y|y\in \mathcal{Y}^s)}\\
	&\leq \mathbb{E}_{y\in \mathcal{Y}^s}\mathbb{E}_{\mathbf{x}\sim p(\mathbf{x})}\frac{p(\mathcal{Y}^s)p(y|y\in \mathcal{Y}^s)}{q(\mathcal{C}_{out}=y|\mathbf{x})p(y|\mathbf{x})}.
	\end{aligned}
    \label{eq7}
	\end{equation}
    
	Using the same inequality on $\mathcal{A}^U$, we have the lower bound of $\mathcal{A}^H$:
    \vspace{-1.ex}
	\begin{equation}
	 \resizebox{0.91\linewidth}{!}{$
	\displaystyle
	\mathcal{A}^H\ge 2/\mathbb{E}_{y\in \mathcal{Y}^s\cup\mathcal{Y}^u}\mathbb{E}_{\mathbf{x}\sim p(\mathbf{x})}\frac{|\mathcal{Y}^s\cup\mathcal{Y}^u|p(\mathcal{Y})p(y|y\in \mathcal{Y})}{|\mathcal{Y}|q(\mathcal{C}_{out}=y|\mathbf{x})p(y|\mathbf{x})},
	  $}
    \label{eq8}
	\end{equation}
    where $\mathcal{Y}$ is $\mathcal{Y}^s$ ($\mathcal{Y}^u$) when $y$ belongs to seen (unseen) classes. To simplify the symbols, we designate $|\mathcal{Y}^s\cup\mathcal{Y}^u|p(\mathcal{Y})/|\mathcal{Y}|$ as $p_0(\mathcal{Y})$ in the following.
    
\begin{figure}[t]
		\centering
		\subfigure
		{
			\includegraphics[width=0.41\textwidth]{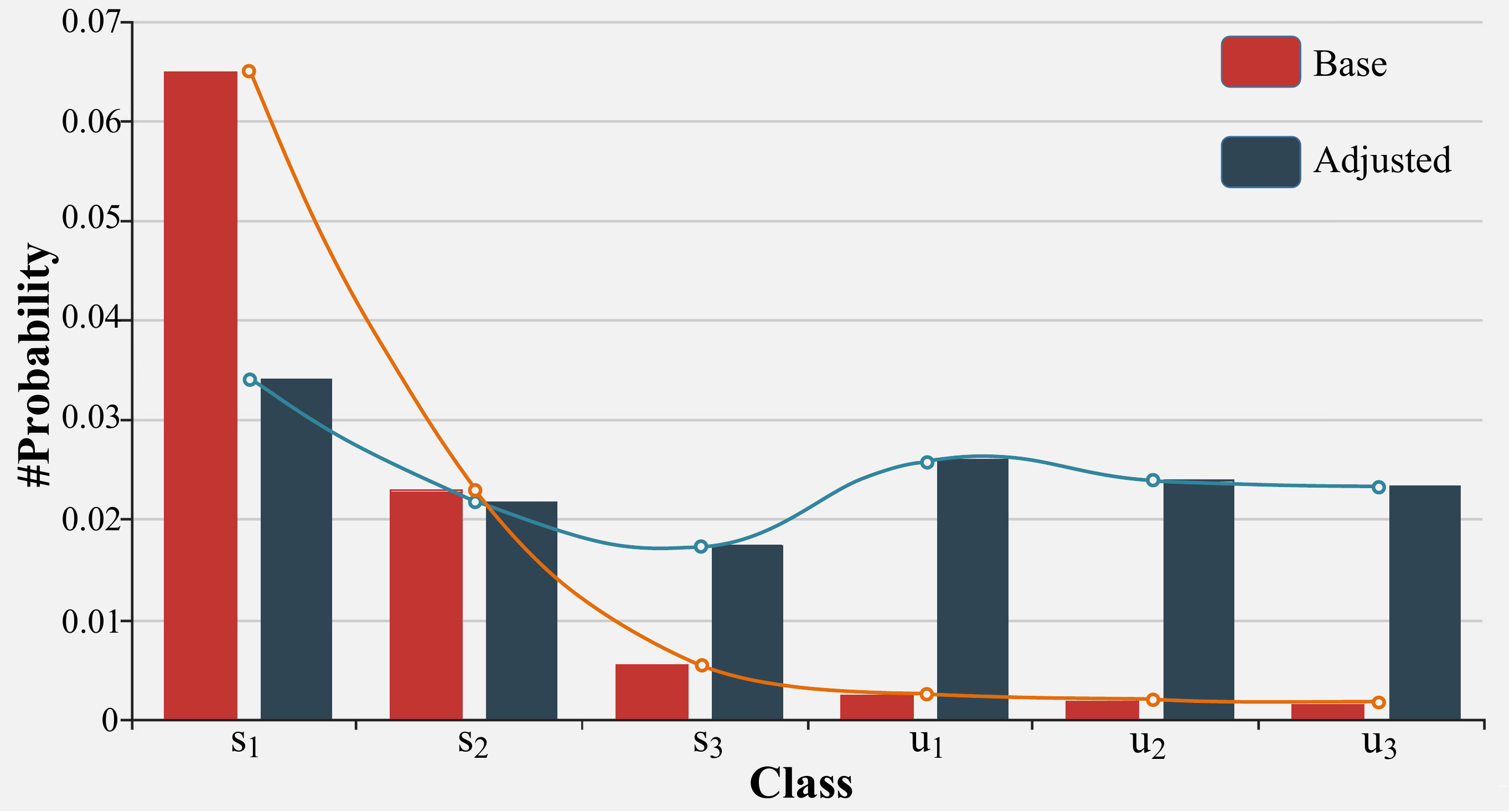}
		}
		\vspace{-1.5ex}
		\caption{A seen (s) and unseen (u) class-prediction-probability example of modeling the base and the adjusted posteriors. The unseen class probabilities are suppressed when modeling the base posterior $p(y|\mathbf{x})$, while the adjusted posterior $p(y|\mathbf{x})/[p_0(\mathcal{Y})p(y|y \in \mathcal{Y})]$ provides a more balanced distribution.}
        
		\vspace{-1.6ex}
		\label{fig2}
	\end{figure}

	Despite the difficulty in determining the Bayesian optimal of $\mathcal{A}^H$, maximizing its lower bound achieves an approximate effect, which is equivalent to minimizing its reciprocal. Intuitively, the denominator term of Eq. \ref{eq8} is minimized if
	\begin{equation}
	\resizebox{0.89\linewidth}{!}{$
		\displaystyle
		q(\mathcal{C}_{out}=y|\mathbf{x})=
		\left\{
		\begin{array}{lr}
		1,~\text{if}~y=\argmax_i\frac{p(i|\mathbf{x})}{p_0(\mathcal{Y})p(i|i \in \mathcal{Y})}\\
		0,~\text{otherwise}\\
		\end{array}
		\right.
		$}
        \label{eq9}
	\end{equation}
	for each $\mathbf{x}$ in $p(\mathbf{x})$. So, given a datum $(\mathbf{x},y_\mathbf{x})$, we change the modeling objective in Eq. \ref{eq1} to the adjusted posterior probability, i.e.,
       \vspace{-0.3ex}
	\begin{equation}
	\displaystyle
	\frac{p(y_\mathbf{x}|\mathbf{x})}{p_0(\mathcal{Y})p(y_\mathbf{x}|y_\mathbf{x} \in \mathcal{Y})},
    \label{eq10}
	\end{equation}
where $p_0(\mathcal{Y})$ refers to the seen-unseen prior (Sec. \ref{prior}) which reflects the bias and homogeneity of pseudo unseen samples. Eq. \ref{eq10} theoretically gives a more balanced predicted probability distribution than the base posterior, as shown in Figure \ref{fig2}. Next, we will show the estimation of the adjusted posterior.
   \vspace{-1.9ex}
	\subsection{ZLA-Based Classifier}
\paragraph{Adjusted Cross-Entropy}
The base posterior probability in Eq. \ref{eq1} is typically estimated with the cross-entropy loss, i.e.,
	\begin{equation}
	\displaystyle
	\mathcal{L}_\mathrm{CE}=\mathrm{log}\sum_{y' \ne y}[1+ \mathrm {exp}(\mathcal{C}_{y'}(x)-\mathcal{C}_y(x))].
    \label{eq11}
	\end{equation}
    
	Referring to researches on class imbalance \cite{tan2020equalization,menon2020long}, we directly model Eq. \ref{eq10} with $\mathcal{C}(\cdot)$ and the posterior becomes
	\begin{equation}
	\displaystyle
	p(y|\mathbf{x}):=p_0(\mathcal{Y})p(y|y \in \mathcal{Y})\cdot\mathrm{softmax}[\mathcal{C}(\mathbf{x})].
    \label{eq12}
	\end{equation}
    
	This enables integrating the conditional class prior $p(y|y\in \mathcal{Y})$ and the seen-unseen prior $p_0(\mathcal{Y})$ into the softmax cross-entropy in a logit adjustment manner. Then the weights in Eq. \ref{eq2} are replaced with
	\begin{equation}
	\displaystyle
	\delta(y,y'):=\frac{p_0(\mathcal{Y}')p(y'|y'\in \mathcal{Y'})}{p_0(\mathcal{Y})p(y|y\in \mathcal{Y})}.
    \label{eq13}
	\end{equation}
    
	The final adjusted cross-entropy is
	\begin{equation}
	\resizebox{0.89\linewidth}{!}{$
		\displaystyle
		\mathcal{L}_{ZLA}=\log [1+\sum_{y' \ne y}\frac{p_0(\mathcal{Y}')p(y'|y'\in \mathcal{Y'})}{p_0(\mathcal{Y})p(y|y\in \mathcal{Y})}  \mathrm {exp}(\mathcal{C}_{y'}(\mathbf{x})-\mathcal{C}_y(\mathbf{x}))].
		$}
        \label{eq14}
	\end{equation}
    
In contrast to the standard cross-entropy form, we consider the specific prior information in the generation-based GZSL setting, which contributes to the balancing results across classes. In practice, we make $p_0(\mathcal{Y}^s)$ much bigger than $p_0(\mathcal{Y}^u)$. This is intuitively explainable from two perspectives: first, small $p_0(\mathcal{Y}^u)/{p_0(\mathcal{Y}^s)}$ promotes seen samples to focus on learning decision boundaries between seen classes; and second, large $p_0(\mathcal{Y}^s)/{p_0(\mathcal{Y}^u)}$ encourages large prediction probabilities for unseen classes, which serves the same purpose as a large generation number of unseen samples \cite{xian2018feature,han2021contrastive}.
	
\paragraph{Semantic Prior Inclusion}
Embedding-based methods \cite{li2019rethinking,skorokhodov2020class} work by learning a semantic-visual direct link. In this case, an extra generator is hard to aid in embedder learning (see Sec. \ref{sec4.5} for experimental results) because the overlap between the knowledge (i.e., semantic-visual link) learned by the generator and the embedder results in semantic information crucial for knowledge transfer not being fully exploited. The proposed ZLA, in contrast, allows for supporting the learning of the semantic-based classifier through an adjustment mechanism. No longer teaching the net the semantic-visual link of unseen classes, the pseudo unseen samples adjust the decision boundaries between seen and unseen classes by weighting the logits, avoiding knowledge overlapping to an extent. Specifically, we adopt a prototype learner $\mathcal{P}$ \cite{li2019rethinking,skorokhodov2020class} to directly map semantics to visual prototypes, and then the adjusted posterior probability of a datum $\mathbf{x}$ is estimated through cosine similarity, i.e.,
\vspace{-1ex}
	\begin{equation}
	\displaystyle
    \begin{aligned}
    \mathcal{C}(\mathbf{x})&:=\cos(\mathbf{x},\mathcal{P}(\mathbf{a}))/\tau  ,
    \end{aligned}
    \label{eq15}
	\end{equation}
	where $\tau$ is the temperature \cite{hinton2015distilling}. In the test phase, the prediction class $y^*$ corresponds to the prototype that achieves the maximum similarity.
	\begin{equation}
	\resizebox{0.89\linewidth}{!}{$
	\displaystyle
	y^*=\argmax_i\frac{p(i|\mathbf{x})}{p_0(\mathcal{Y})p(i|i \in \mathcal{Y})}:=\argmax_i \cos(\mathbf{x},\mathcal{P}(\mathbf{a}_i)).
	 $}
    \label{eq16}
	\end{equation}
	\section{Experiments}
\begin{table*}[t]
		\centering
		\resizebox{\textwidth}{!}{
			\begin{tabular}{@{}cllcccccccccccc}
				\toprule
		&\multirow{2}{*}{Method} & \multirow{2}{*}{Reference}  & \multicolumn{3}{c}{AWA2}            & \multicolumn{3}{c}{CUB}                                                  & \multicolumn{3}{c}{SUN}                                                  & \multicolumn{3}{c}{APY}                                                  \\
				&            &                & $\mathcal{A}^U$      & $\mathcal{A}^S$      & $\mathcal{A}^H$     & $\mathcal{A}^U$                        & \multicolumn{1}{c}{$\mathcal{A}^S$} & \multicolumn{1}{c}{$\mathcal{A}^H$} & $\mathcal{A}^U$                        & \multicolumn{1}{c}{$\mathcal{A}^S$} & \multicolumn{1}{c}{$\mathcal{A}^H$} & $\mathcal{A}^U$                        & \multicolumn{1}{c}{$\mathcal{A}^S$} & \multicolumn{1}{c}{$\mathcal{A}^H$} \\
				\midrule

\multicolumn{1}{c|}{\multirow{4}{*}{$\dagger$}} 			&Li et al.               & ICCV \cite{li2019rethinking}            & 56.4          & \underline{81.4} & 66.7          & 47.4          & 47.6          & 47.5          & 36.3          & {\ul 42.8}    & 39.3          & 26.5          & \textbf{74.0}    & 39.0          \\
   \multicolumn{1}{c|}{}		&DVBE                   &CVPR \cite{xu2020attribute}    &63.6 &70.8 &67.0 &53.2 &60.2 &56.5  &45.0 &37.2 &40.7  &32.6 &58.3 &41.8\\
   \multicolumn{1}{c|}{}	&RGEN &ECCV\cite{xie2020region} &{\bf 67.1} &76.5 &71.5  &60.0 &{\bf 73.5} &66.1 &44.0 &31.7 &36.8 &30.4 &48.1 &37.2\\
	\multicolumn{1}{c|}{}		&APN                   & NeurIPS \cite{min2020domain}                       &56.5 &78.0 &65.5        & 65.3 &\underline{69.3} &67.2     &41.9 &34.0 &37.6     & -             & -             & -             \\
\midrule
\midrule
\multicolumn{1}{c|}{\multirow{9}{*}{$\ddagger$}}  	&Li et al.               & AAAI \cite{li2021generalized}            & 56.9          & 80.2    & 66.6          & 51.1          & 58.2          & 54.4          & 47.6          & 36.6          & 41.4          & -             & -             & -             \\
\multicolumn{1}{c|}{}        &GCM-CF   &CVPR \cite{yue2021counterfactual}        &60.4 &75.1 &67.0  &61.0 &59.7 &60.3    &47.9 &37.8 &\underline{42.2}   &37.1 &56.8 &44.9\\
 \multicolumn{1}{c|}{}		&AGZSL                   & ICLR \cite{chou2020adaptive}                       & 65.1          & 78.9          &71.3          & 41.4          & 49.7          &45.2         & 29.9           & {\bf 40.2}           &34.3          &35.1 &65.5 &45.7  \\
	\multicolumn{1}{c|}{}			&FREE                    & ICCV \cite{chen2021free}              & 60.4          & 75.4          & 67.1          & 55.7          & 59.9          & 57.7          & 47.4          & 37.2          & 41.7          & -             & -             & -             \\
\multicolumn{1}{c|}{}		&SDGZSL                  & ICCV \cite{chen2021semantics}        & 64.6 & 73.6 & 68.8                     & 59.9                     & 66.4                     & 63.0                     & -                        & -                        & -                        & 38.0                     & 57.4                     & 45.7                     \\
\noalign{\smallskip}
\cline{2-15}
\noalign{\smallskip}
  \multicolumn{1}{c|}{}             & f-CLSWGAN               & CVPR \cite{xian2018feature}             & 57.7 & 71.0 & 63.7            & 59.4          & 63.3         & 61.3         & 46.2          & 35.2          & 40.0          & 32.5             & 57.2             & 41.5             \\
	\multicolumn{1}{c|}{}			&WGAN+{\bf ZLAP}                & {\bf Proposed}                    & \underline{65.4} & {\bf82.2} & {\bf72.8}                  & \multicolumn{1}{l}{{\bf 73.0}} & 64.8                 & {\bf 68.7}                  & \multicolumn{1}{l}{50.1} & \underline{38.0}                  &{\bf 43.2}                  & \multicolumn{1}{l}{{\bf 40.2}} & 53.8                  &\underline{ 46.0}                 \\
    \noalign{\smallskip}
\cline{2-15}
\noalign{\smallskip}
    \multicolumn{1}{c|}{}            &CE-GZSL                 & CVPR \cite{han2021contrastive}     & 65.3          & 75.0         & 69.9          & 66.9    & 65.9          & 66.4          & {\bf 52.4}          & 34.3          & 41.5          & 28.3            & \underline{65.8}             &39.6             \\
     \multicolumn{1}{c|}{}           &CE-GZSL+{\bf ZLAP}    &{\bf Proposed}   &64.8 &80.9 &\underline{72.0}   &\underline{71.2} &66.2 &\underline{68.6} &\underline{50.9}&35.7&42.0   &\underline{38.3} &60.9 &{\bf47.0}\\
				\bottomrule
			\end{tabular}
		}
		\caption{GZSL performance comparisons with state-of-the-art methods. $\mathcal{A}^U$ and $\mathcal{A}^S$ denote the per-class accuracy (\%) on unseen and seen classes, respectively, and $\mathcal{A}^H$ is their harmonic mean. The best results are bolded, and the underlines indicate the second-place results. $\dagger$ and $\ddagger$ represent whether a generator is employed to obtain the pseudo unseen samples, respectively ($\ddagger$ indicates yes, and $\dagger$ is the opposite). {\bf ZLAP} is the proposed zero-shot logit adjustment prototype learner.}
        \vspace{-2ex}
		\label{tab2}
	\end{table*}
	\subsection{Datasets and Metrics}
\label{metric}
\paragraph{Benchmark Datasets}
We study GZSL performed in Animals with Attributes 2 (AWA2) \cite{lampert2013attribute}, Attribute Pascal and Yahoo (APY) \cite{farhadi2009describing}, Caltech-UCSD Birds-200-2011 (CUB) \cite{wah2011caltech}, and SUN Attribute (SUN) \cite{patterson2012sun}, following the common split (version 2) proposed in \cite{xian2017zero}. AWA2 includes 50 animal species and 85 attribute annotations, accounting 37,322 samples. APY contains 32 classes of 15,339 samples and 64 attributes. CUB consists of 11,788 samples with 200 bird species, annotated by 312 attributes. SUN carries 14,340 images from 717 different scenario-style with 102 attributes.
	
\paragraph{Visual Representations and Semantic Descriptors}
We follow \cite{xian2017zero} to represent images as the 2048-D ResNet-101 \cite{he2016deep} features. Moreover, we regard the artificial attribute annotations that come with the datasets as the semantic descriptors of AWA2, APY, and SUN, and the 1024-dimensional character-based CNN-RNN features \cite{reed2016learning} generated from textual descriptions as the semantics of CUB.
	
\paragraph{Evaluation Protocol}
We calculate the average per-class top-1 accuracy for unseen and seen classes, respectively, denoted as $\mathcal{A}^U$ and $\mathcal{A}^S$. Then the metric $\mathcal{A}^H$ for GZSL is represented as their harmonic mean.\cite{xian2017zero}.
	
	\subsection{Implementation Details}
	Since our work focuses on studying a plug-in classifier, we test it on WGAN \cite{gulrajani2017improved}, with the same generator and discriminator structures as \cite{xian2018feature}. Our prototype learner $\mathcal{P}$ consists of a multi-layer perceptron (MLP) with a single 1024-D hidden layer activated by LeakyReLU and no activation function in the output. The Adam optimizer is employed with a learning rate of $1\times10^{-3}$, and the batch size is set at 512 for evaluating our design. When plugging into CE-GZSL \cite{han2021contrastive}, we employ a batch size of 256 in SUN and 512 in other datasets instead of the default 4096 (due to device limitations) while maintaining all other settings in the published paper.
	\subsection{Comparison with State-of-the-Arts}
We apply the proposed classifier to the vanilla WGAN (compare to f-CLSWGAN \cite{xian2018feature}) and the more advanced CE-GZSL \cite{han2021contrastive} to show its effect and compatibility in different generative frameworks. The baseline results are obtained from the official codes. As shown in Table \ref{tab2}, the combination of the proposed classifier and the basic generative framework (WGAN) outperforms SoTAs, demonstrating the excellent class balancing capability of ZLA. We note that the performance gain provided by our module is not as significant in the highly fine-grained SUN dataset as it is in other datasets. It is mainly due to the minor bias problem of generated unseen samples in the case of multiple classes and modest feature variations that it takes the limited strengths of our approach. Moreover, our classifier improves the performance of both f-CLSWGAN and CE-GZSL, even though CE-GZSL has already produced decent results, proving its plug-in ability in two-stage frameworks.
       \begin{figure}[t]
		\centering
		\subfigure 
		{
			\includegraphics[width=0.227\textwidth]{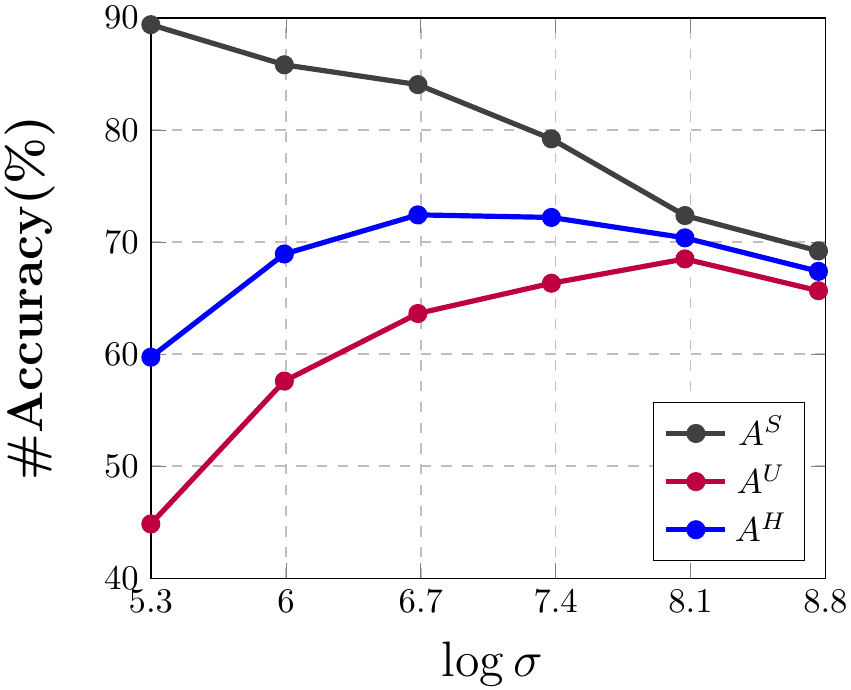}
		}
        \subfigure 
		{
			\includegraphics[width=0.2297\textwidth]{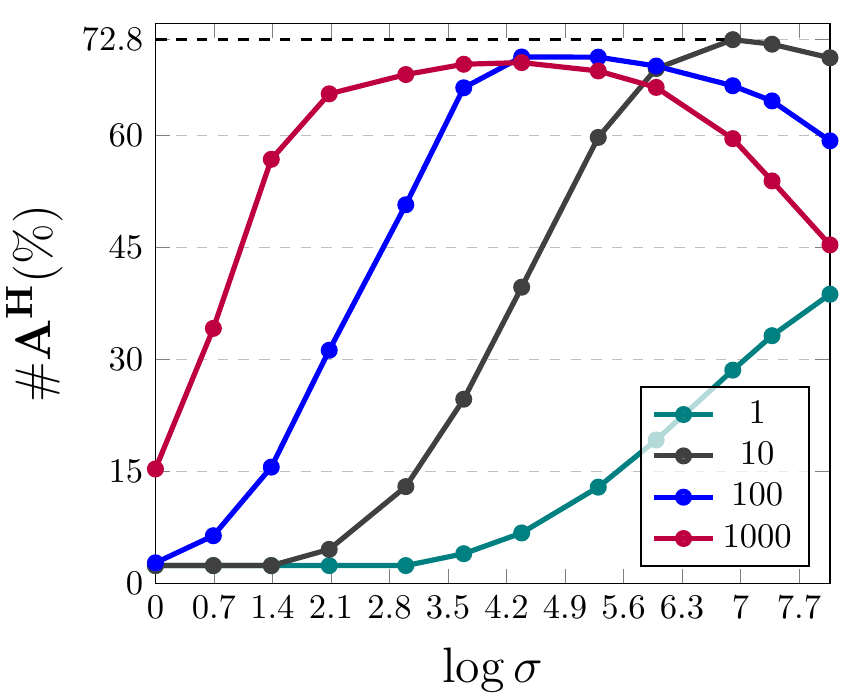}
		}
		\vspace{-2.9ex}
		\caption{{\bf Left:} hyperparemeter analysis of $\sigma$ (see \ref{hype} for its definition) on AWA2. {\bf Right:} $\log\sigma$ varies w.r.t. generation number per unseen class. Large generation number lowers the performance cap (72.8 with 10 generated vs. 69.7 with 1000 generated).}
		\vspace{-2ex}
		\label{fig3}
	\end{figure}
\subsection{Hyperparameters}
\label{hype}
Three key factors are involved in our work, i.e., the generation number (per unseen class) $N_g$, $p_0(\mathcal{Y})$ in Eq. \ref{eq10}, and the temperature $\tau$ in Eq. \ref{eq15}. Following \cite{skorokhodov2020class}, $\tau$ is fixed at 0.04 for all experiments, since it has a slight bearing on our study. We first examine $p_0(\mathcal{Y})$ in the form of the ratio of $p_0(\mathcal{Y}^s)$ to $p_0(\mathcal{Y}^u)$, which more intuitively reflects the seen-unseen dichotomy. The ratio is denoted as $\sigma$, and its effect is plotted in Figure \ref{fig3} (left), where a reverse variation of seen and unseen accuracy can be observed, demonstrating ZLA's capacity to moderate the prediction between classes. Figure \ref{fig3} (right) depicts the influence carried by $N_g$. Although $N_g$ also posses the ability to affect the accuracy, a large generation number lowers the performance cap (72.8 with 10 generated vs. 69.7 with 1000 generated), indicating the harm of re-sampling the biased samples. In the major experiment (Table \ref{tab2}), we generate 10 samples per unseen class for all datasets to contrast with f-CLSWGAN \cite{xian2018feature}, and the best results are obtained when $\sigma$ is set to 1000, 30, 60, and 300 for AWA2, CUB, SUN, and APY, respectively (better results are possible by trading off between $N_g$ and $\sigma$). In comparison to CE-GZSL \cite{han2021contrastive}, we keep the published generated number and take the value of $\sigma$ as 28, 9, 7, and 920 for the above datasets.
	\subsection{Ablation Study}
	\label{sec4.5}
	In this section, we perform ablation studies to validate our design and display the key elements in our implementation.
	\begin{table}[t]
		\centering
		\resizebox{0.482\textwidth}{!}{
			\begin{tabular}{lccccccc}
				\toprule
				\multirow{2}{*}{Method} & \multirow{2}{*}{ReLU}     & \multicolumn{3}{c}{AWA2} & \multicolumn{3}{c}{CUB} \\
				&                           & $\mathcal{A}^U$      &$\mathcal{A}^S$      & $\mathcal{A}^H$      & $\mathcal{A}^U$      & $\mathcal{A}^S$      &$\mathcal{A}^H$     \\ 
				\midrule    
				\multirow{2}{*}{Non Gen.}           & $\checkmark$ & 55.1   & 82.0   & 65.9   & 68.3   & 54.8   & 60.8  \\
				& $\times$     & 24.6   & {\bf 89.9}   & 38.6   & 60.5   & {\bf 66.6}   & 63.4 \\
				\midrule    
				\multirow{2}{*}{Gen.}     &$\checkmark$ & 56.5   & 81.3   & 66.7   & 64.7   & 64.7   & 64.7  \\
				& $\times$     &{\bf 65.4}   & 82.2   & {\bf 72.8}    &{\bf 73.0} & 64.8 &{\bf 68.7}  \\ 
				\bottomrule
			\end{tabular}
		}
        	\vspace{-0.5ex}
		\caption{Comparison of the pure prototype learner and the generation-based ZLA prototype learner. {\bf Non Gen.:} a latest proposed pure prototype learner, implemented by the official code (with the post-hoc correction removed for a fair comparison). {\bf Gen.:} WGAN-based zero-shot logit adjustment prototype learner.}
        	\vspace{-0.5ex}
		\label{tab3}
	\end{table}
    
    	\begin{table}[t]
		\centering
        \resizebox{0.482\textwidth}{!}{
		\begin{tabular}{lcccccccc}
			\toprule
			\multirow{2}{*}{Classifier}& \multirow{2}{*}{ZLA} &\multirow{2}{*}{L.N.} & \multicolumn{3}{c}{AWA2} & \multicolumn{3}{c}{CUB} \\
			&&& $\mathcal{A}^U$      & $\mathcal{A}^S$      & $\mathcal{A}^H$      & $\mathcal{A}^U$      & $\mathcal{A}^S$      &$\mathcal{A}^H$     \\
			\midrule
			\multirow{3}{*}{Vanilla}    &$\checkmark$  &$\times$ & 40.2   & {\bf 82.5}   & 54.1   & 61.4   & 52.3   & 56.5  \\
                         &$\times$    &$\checkmark$& 57.7 & 71.0 & 63.7 &59.4&63.3&61.3 \\
                           &$\checkmark$    &$\checkmark$&61.2&74.6&67.3&66.8 &63.5 &65.1 \\
                         \midrule
			\multirow{3}{*}{Proto.}         &$\checkmark$   &$\times$   &{\bf 65.4}   & 82.2   & {\bf72.8}   &{\bf 73.0} & {\bf 64.8} &{\bf 68.7} \\
                         &$\times$ &$\checkmark$ &50.7 &75.8 &60.8 &60.0 &63.4 &61.4\\
                          &$\checkmark$    &$\checkmark$&64.1&73.1&68.3 &72.4&63.0&67.4 \\
			\bottomrule
		\end{tabular}
             }
        	\vspace{-0.5ex}
		\caption{\textbf{Vanilla softmax classifier} vs. \textbf{prototype learner}, based on WGAN, where L.N. represents a large generation number.}
        	\vspace{-2ex}  
		\label{tab4}
	\end{table}

\paragraph{Function of ZLA}
Figure \ref{fig3} (right) shows the function of ZLA, where $\log \sigma=0$ means $p_0(\mathcal{Y}^s)=p_0(\mathcal{Y}^u)$, i.e., without major adjustments. The effect of ZLA is reflected in the comparison of different $\sigma$ values. Intuitively, an adequate $\sigma$ value produces considerable performance gains in varying generation numbers, especially when the number is minimal. Furthermore, ZLA enables the prototype learner to be effective in generative scenarios, as illustrated in Table \ref{tab4} (second lines in two baselines reflect previous ineffectiveness), which further reduces the reliance on the generation number and thereby resolving the previously discussed bias problem.
    
\paragraph{Beyond the Standard Prototype Learner}
Existing prototype leaners \cite{li2019rethinking,skorokhodov2020class} are simply established on the semantic-visual links of seen classes, generalizing to unseen classes based on semantic similarities. In this paper, we find the last ReLU layer is crucial to these approaches' hitherto unseen class performances. To investigate the effect of the ReLU function, we compare the latest pure prototype learner \cite{skorokhodov2020class} with the WGAN-based ZLAP in Table \ref{tab3}. When the ReLU layer is removed, as shown in Table \ref{tab3}, the seen accuracy improves in both baselines, whereas the unseen accuracy decreases if pseudo unseen samples are not accessible. Zeroing the negative output layer values intuitively affects (seen class) prototype expression. However, it provides a regularization which narrows the gap between the unseen class prototypes and the instances when (pseudo) unseen instances are unavailable in training, since the visual feature components are likewise larger than or equal to zero \cite{xian2017zero}. In this sense, our proposed ZLA allows the model to remove the ReLU layer by adjusting the unseen class expression using the synthetic unseen instances, resulting in a win-win situation for both seen and unseen class accuracy.

\paragraph{Beyond the Vanilla Softmax Classifier}
In Table \ref{tab4}, we compare the prototype-based classifier to the vanilla softmax classifier to validate its necessity. Results reveal that the prototype learner beats the vanilla softmax classifier thoroughly when ZLA is employed. The explanations are as follows: (1) the prototype learner enables further regularization on unseen class prototypes by learning semantic-prototype relations with real-world data, whereas the vanilla softmax classifier learns to distinguish unseen classes solely by generated samples; and (2) classifier weights mapped from semantics focus more on category-distinctive information, i.e., semantic information. Specifically, in coarse-grained datasets like AWA2, the generated unseen class samples meet a more serious bias problem, causing the classifier to incorrectly detect unseen classes. The prototype learner, on the other hand, can provide supplemental information to unseen class weights by comprehending the semantic-visual links in seen classes. In fine-grained datasets such as CUB, samples from different classes are relatively close together, making it challenging to separate them correctly. The semantics-based classifier, which contains intrinsically category-distinctive information, aids in increased discrimination.

\subsection{Time Complexity Analysis}
We note that some recent generation-based methods \cite{han2021contrastive,chen2021semantics} also mine the semantic discriminant information of samples. However, these methods are typically built on class contrast during generator training, which yields a time complexity of $O(N|\mathcal{Y}^s|)$ in the training phase ($N$ is the data size). In comparison, the proposed classifier combined with the vanilla WGAN achieves a comparable result with $O(N)$ time complexity. Moreover, the proposed method allows for a much smaller (10 vs. 4000 in AWA2) synthetic number in the classifier training phase, resulting in further time savings.
\section{Conclusion}
In this work, we theoretically include the logit adjustment tech in the generation-based GZSL. We begin by examining the bias and homogeneity of the generated unseen samples, which build the seen-unseen prior. Then we derive an adjusted posterior from the seen-unseen balanced metric, which enables integrating the seen-unseen prior into the original cross-entropy via logit adjustment. Considering the zero-shot setting, we call our approach Zero-Shot Logit Adjustment. Based on ZLA, we inject the semantic information into the classifier, which always fails in existing two-stage methods.

Our work explores the underutilized potential of the generation-based GZSL by breaking the previous inconsistency between the classifier's training objectives and testing metrics. This approach allows for greatly fewer generated unseen samples, achieving SoTA with little time consumption.

{\small\bibliographystyle{named}
	\bibliography{ijcai22}}

\begin{thebibliography}{}

\bibitem[\protect\citeauthoryear{Buda \bgroup \em et al.\egroup
  }{2018}]{buda2018systematic}
Mateusz Buda, Atsuto Maki, and Maciej~A Mazurowski.
\newblock A systematic study of the class imbalance problem in convolutional
  neural networks.
\newblock {\em Neural Networks}, 106:249--259, 2018.

\bibitem[\protect\citeauthoryear{Cao \bgroup \em et al.\egroup
  }{2019}]{cao2019learning}
Kaidi Cao, Colin Wei, Adrien Gaidon, Nikos Arechiga, and Tengyu Ma.
\newblock Learning imbalanced datasets with label-distribution-aware margin
  loss.
\newblock {\em NeurIPS}, 2019.

\bibitem[\protect\citeauthoryear{Chao \bgroup \em et al.\egroup
  }{2016}]{chao2016empirical}
Wei-Lun Chao, Soravit Changpinyo, Boqing Gong, and Fei Sha.
\newblock An empirical study and analysis of generalized zero-shot learning for
  object recognition in the wild.
\newblock In {\em ECCV}, pages 52--68, 2016.

\bibitem[\protect\citeauthoryear{Chawla \bgroup \em et al.\egroup
  }{2002}]{chawla2002smote}
Nitesh~V Chawla, Kevin~W Bowyer, Lawrence~O Hall, and W~Philip Kegelmeyer.
\newblock Smote: synthetic minority over-sampling technique.
\newblock {\em JAIR}, 16:321--357, 2002.

\bibitem[\protect\citeauthoryear{Chen \bgroup \em et al.\egroup
  }{2021a}]{chen2021free}
Shiming Chen, Wenjie Wang, Beihao Xia, Qinmu Peng, Xinge You, Feng Zheng, and
  Ling Shao.
\newblock Free: Feature refinement for generalized zero-shot learning.
\newblock {\em ICCV}, 2021.

\bibitem[\protect\citeauthoryear{Chen \bgroup \em et al.\egroup
  }{2021b}]{chen2021semantics}
Zhi Chen, Yadan Luo, Ruihong Qiu, Zi~Huang, Jingjing Li, and Zheng Zhang.
\newblock Semantics disentangling for generalized zero-shot learning.
\newblock In {\em ICCV}, 2021.

\bibitem[\protect\citeauthoryear{Chou \bgroup \em et al.\egroup
  }{2021}]{chou2020adaptive}
Yu-Ying Chou, Hsuan-Tien Lin, and Tyng-Luh Liu.
\newblock Adaptive and generative zero-shot learning.
\newblock In {\em ICLR}, 2021.

\bibitem[\protect\citeauthoryear{Collell \bgroup \em et al.\egroup
  }{2016}]{collell2016reviving}
Guillem Collell, Drazen Prelec, and Kaustubh Patil.
\newblock Reviving threshold-moving: a simple plug-in bagging ensemble for
  binary and multiclass imbalanced data.
\newblock {\em CoRR}, 2016.

\bibitem[\protect\citeauthoryear{Farhadi \bgroup \em et al.\egroup
  }{2009}]{farhadi2009describing}
Ali Farhadi, Ian Endres, Derek Hoiem, and David Forsyth.
\newblock Describing objects by their attributes.
\newblock In {\em CVPR}, pages 1778--1785, 2009.

\bibitem[\protect\citeauthoryear{Fu \bgroup \em et al.\egroup
  }{2014}]{fu2014transductive}
Yanwei Fu, Timothy~M Hospedales, Tao Xiang, Zhenyong Fu, and Shaogang Gong.
\newblock Transductive multi-view embedding for zero-shot recognition and
  annotation.
\newblock In {\em ECCV}, pages 584--599. Springer, 2014.

\bibitem[\protect\citeauthoryear{Goodfellow \bgroup \em et al.\egroup
  }{2014}]{goodfellow2014generative}
Ian Goodfellow, Jean Pouget-Abadie, Mehdi Mirza, Bing Xu, David Warde-Farley,
  Sherjil Ozair, Aaron Courville, and Yoshua Bengio.
\newblock Generative adversarial nets.
\newblock {\em NeurIPS}, 27, 2014.

\bibitem[\protect\citeauthoryear{Gulrajani \bgroup \em et al.\egroup
  }{2017}]{gulrajani2017improved}
Ishaan Gulrajani, Faruk Ahmed, Martin Arjovsky, Vincent Dumoulin, and Aaron
  Courville.
\newblock Improved training of wasserstein gans.
\newblock {\em arXiv preprint arXiv:1704.00028}, 2017.

\bibitem[\protect\citeauthoryear{Han \bgroup \em et al.\egroup
  }{2021}]{han2021contrastive}
Zongyan Han, Zhenyong Fu, Shuo Chen, and Jian Yang.
\newblock Contrastive embedding for generalized zero-shot learning.
\newblock In {\em CVPR}, pages 2371--2381, 2021.

\bibitem[\protect\citeauthoryear{He \bgroup \em et al.\egroup
  }{2016}]{he2016deep}
Kaiming He, Xiangyu Zhang, Shaoqing Ren, and Jian Sun.
\newblock Deep residual learning for image recognition.
\newblock In {\em CVPR}, pages 770--778, 2016.

\bibitem[\protect\citeauthoryear{Hinton \bgroup \em et al.\egroup
  }{2015}]{hinton2015distilling}
Geoffrey Hinton, Oriol Vinyals, and Jeff Dean.
\newblock Distilling the knowledge in a neural network.
\newblock {\em NeurIPS}, 2015.

\bibitem[\protect\citeauthoryear{Kingma and Welling}{2013}]{kingma2013auto}
Diederik~P Kingma and Max Welling.
\newblock Auto-encoding variational bayes.
\newblock {\em ICLR}, 2013.

\bibitem[\protect\citeauthoryear{Lampert \bgroup \em et al.\egroup
  }{2009}]{lampert2009learning}
Christoph~H Lampert, Hannes Nickisch, and Stefan Harmeling.
\newblock Learning to detect unseen object classes by between-class attribute
  transfer.
\newblock In {\em CVPR}, pages 951--958, 2009.

\bibitem[\protect\citeauthoryear{Lampert \bgroup \em et al.\egroup
  }{2013}]{lampert2013attribute}
Christoph~H Lampert, Hannes Nickisch, and Stefan Harmeling.
\newblock Attribute-based classification for zero-shot visual object
  categorization.
\newblock {\em IEEE TPAMI}, 36(3):453--465, 2013.

\bibitem[\protect\citeauthoryear{Li \bgroup \em et al.\egroup
  }{2019}]{li2019rethinking}
Kai Li, Martin~Renqiang Min, and Yun Fu.
\newblock Rethinking zero-shot learning: A conditional visual classification
  perspective.
\newblock In {\em ICCV}, pages 3583--3592, 2019.

\bibitem[\protect\citeauthoryear{Li \bgroup \em et al.\egroup
  }{2021}]{li2021generalized}
Xiangyu Li, Zhe Xu, Kun Wei, and Cheng Deng.
\newblock Generalized zero-shot learning via disentangled representation.
\newblock In {\em AAAI}, volume~35, pages 1966--1974, 2021.

\bibitem[\protect\citeauthoryear{Lin \bgroup \em et al.\egroup
  }{2017}]{lin2017focal}
Tsung-Yi Lin, Priya Goyal, Ross Girshick, Kaiming He, and Piotr Doll{\'a}r.
\newblock Focal loss for dense object detection.
\newblock In {\em ICCV}, pages 2980--2988, 2017.

\bibitem[\protect\citeauthoryear{Liu \bgroup \em et al.\egroup
  }{2018}]{liu2018generalized}
Shichen Liu, Mingsheng Long, Jianmin Wang, and Michael~I Jordan.
\newblock Generalized zero-shot learning with deep calibration network.
\newblock In {\em NeurIPS}, pages 2005--2015, 2018.

\bibitem[\protect\citeauthoryear{Menon \bgroup \em et al.\egroup
  }{2013}]{menon2013statistical}
Aditya Menon, Harikrishna Narasimhan, Shivani Agarwal, and Sanjay Chawla.
\newblock On the statistical consistency of algorithms for binary
  classification under class imbalance.
\newblock In {\em ICML}, pages 603--611. PMLR, 2013.

\bibitem[\protect\citeauthoryear{Menon \bgroup \em et al.\egroup
  }{2020}]{menon2020long}
Aditya~Krishna Menon, Sadeep Jayasumana, Ankit~Singh Rawat, Himanshu Jain,
  Andreas Veit, and Sanjiv Kumar.
\newblock Long-tail learning via logit adjustment.
\newblock {\em arXiv preprint arXiv:2007.07314}, 2020.

\bibitem[\protect\citeauthoryear{Min \bgroup \em et al.\egroup
  }{2020}]{min2020domain}
Shaobo Min, Hantao Yao, Hongtao Xie, Chaoqun Wang, Zheng-Jun Zha, and Yongdong
  Zhang.
\newblock Domain-aware visual bias eliminating for generalized zero-shot
  learning.
\newblock In {\em CVPR}, pages 12664--12673, 2020.

\bibitem[\protect\citeauthoryear{Patterson and Hays}{2012}]{patterson2012sun}
Genevieve Patterson and James Hays.
\newblock Sun attribute database: Discovering, annotating, and recognizing
  scene attributes.
\newblock In {\em CVPR}, pages 2751--2758, 2012.

\bibitem[\protect\citeauthoryear{Reed \bgroup \em et al.\egroup
  }{2016}]{reed2016learning}
Scott Reed, Zeynep Akata, Honglak Lee, and Bernt Schiele.
\newblock Learning deep representations of fine-grained visual descriptions.
\newblock In {\em CVPR}, pages 49--58, 2016.

\bibitem[\protect\citeauthoryear{Shen \bgroup \em et al.\egroup
  }{2020}]{shen2020invertible}
Yuming Shen, Jie Qin, Lei Huang, Li~Liu, Fan Zhu, and Ling Shao.
\newblock Invertible zero-shot recognition flows.
\newblock In {\em ECCV}, pages 614--631, 2020.

\bibitem[\protect\citeauthoryear{Skorokhodov and
  Elhoseiny}{2021}]{skorokhodov2020class}
Ivan Skorokhodov and Mohamed Elhoseiny.
\newblock Class normalization for (continual)? generalized zero-shot learning.
\newblock {\em ICLR}, 2021.

\bibitem[\protect\citeauthoryear{Tan \bgroup \em et al.\egroup
  }{2020}]{tan2020equalization}
Jingru Tan, Changbao Wang, Buyu Li, Quanquan Li, Wanli Ouyang, Changqing Yin,
  and Junjie Yan.
\newblock Equalization loss for long-tailed object recognition.
\newblock In {\em CVPR}, pages 11662--11671, 2020.

\bibitem[\protect\citeauthoryear{Wah \bgroup \em et al.\egroup
  }{2011}]{wah2011caltech}
Catherine Wah, Steve Branson, Peter Welinder, Pietro Perona, and Serge
  Belongie.
\newblock The caltech-ucsd birds-200-2011 dataset.
\newblock {\em california institute of technology}, 2011.

\bibitem[\protect\citeauthoryear{Xian \bgroup \em et al.\egroup
  }{2017}]{xian2017zero}
Yongqin Xian, Bernt Schiele, and Zeynep Akata.
\newblock Zero-shot learning-the good, the bad and the ugly.
\newblock In {\em CVPR}, pages 4582--4591, 2017.

\bibitem[\protect\citeauthoryear{Xian \bgroup \em et al.\egroup
  }{2018}]{xian2018feature}
Yongqin Xian, Tobias Lorenz, Bernt Schiele, and Zeynep Akata.
\newblock Feature generating networks for zero-shot learning.
\newblock In {\em CVPR}, pages 5542--5551, 2018.

\bibitem[\protect\citeauthoryear{Xie \bgroup \em et al.\egroup
  }{2020}]{xie2020region}
Guo-Sen Xie, Li~Liu, Fan Zhu, Fang Zhao, Zheng Zhang, Yazhou Yao, Jie Qin, and
  Ling Shao.
\newblock Region graph embedding network for zero-shot learning.
\newblock In {\em ECCV}, pages 562--580. Springer, 2020.

\bibitem[\protect\citeauthoryear{Xu \bgroup \em et al.\egroup
  }{2020}]{xu2020attribute}
Wenjia Xu, Yongqin Xian, Jiuniu Wang, Bernt Schiele, and Zeynep Akata.
\newblock Attribute prototype network for zero-shot learning.
\newblock {\em NeurIPS}, 2020.

\bibitem[\protect\citeauthoryear{Yue \bgroup \em et al.\egroup
  }{2021}]{yue2021counterfactual}
Zhongqi Yue, Tan Wang, Qianru Sun, Xian-Sheng Hua, and Hanwang Zhang.
\newblock Counterfactual zero-shot and open-set visual recognition.
\newblock In {\em CVPR}, pages 15404--15414, 2021.

\end{thebibliography}
\end{document}